# Modelling and Reasoning Techniques for Context – Aware Computing in Intelligent Transportation System

M. Swarnamugi [1] and Dr. R. Chinnaiyan [2]

[1] Assistant Professor, Department of MCA, Jyoti Nivas College, Bengaluru.
Research Scholar, Department of MCA, VTU, CMR Institute of Technology, Bengaluru
swathidevan@gmail.com
[2] Associate Professor, Department of Information Science and Engineering, CMR Institute of Technology, Bengaluru.
vijayachinns@gmail.com, chinnaiyan.r@cmrit.ac.in

## Abstract

The emergence of IoT technology and recent advancement in sensor networks enabled transportation systems to a new dimension called Intelligent Transportation System (ITS). Due to increased usage of vehicles and communication among entities in road traffic scenarios, the amount of raw data generation in ITS is huge. This raw data are to be processed to infer contextual information and provide new services related to different modes of road transport such as traffic signal management, accident prediction, object detection etc.,. To understand the importance of context, this article aims to study context-awareness in the Intelligent Transportation System. We present a review on prominent applications developed in the literature concerning context – awareness in the intelligent transportation system. The objective of this survey article is (i) to highlight context and its features in ITS (ii) to address the applicability of modelling techniques and reasoning approaches in ITS (iii) to shed light on impact of IoT and machine learning in ITS development.

**Keywords** – Intelligent transportation system, Context, Context life cycle, Sensors, Machine learning.

## I. Introduction

The Intelligent Transportation System (ITS) plays a significant role in the transformation of conventional transportation by enhancing user convenience and safety, ensuring efficiency and improving transportation networks. Many countries around the globe have widely accepted and started implementing intelligent transportation systems as a solution to the current road traffic management practices [1]. Research in ITS aims to afford services relating to different modes of road transport and assist users to be better informed on road safety and make safer, coordinated, comfort and smarter use of transportation [2].

The traditional road transportation currently in practise encompasses more of human to human communication. Imparting intelligence to it increases many technical challenges in enabling human to machine communication and machine to machine communication. Yet thanks to advanced technologies such as electronic sensor technology, wireless sensor network, mobile computing, cloud computing, Internet of Things, computer vision, data transmission technology and intelligent control have enabled large scale deployment of intelligent transportation in reality [56]. ITS has raised opportunities to explore and investigate problems that exist in road transportation such as destination prediction, demand prediction, object detection, traffic flow control management and prediction, demand prediction, travel time estimation, predicting traffic accident severity, predicting the mode of transportation, navigation, demand serving like optimized shared – taxi systems, parking slot management etc.,[3]. The solution to all these problems enable multiple objects/entities to communicate with each other. Figure 1 depicts the various applications of ITS and the problem that it addresses. The increase in the number of objects connected to ITS, will pave the way for the massive generation of data collected from various sources such as on-board terminals (sensors mounted on vehicles), base stations (sensors mounted on environment. Ex: surveillance camera) web services, cloud and user interface. The generated data will have profound insight on the design and implementation of ITS to make it safer and smarter. This research work aims to describe the profound insight on data as context.

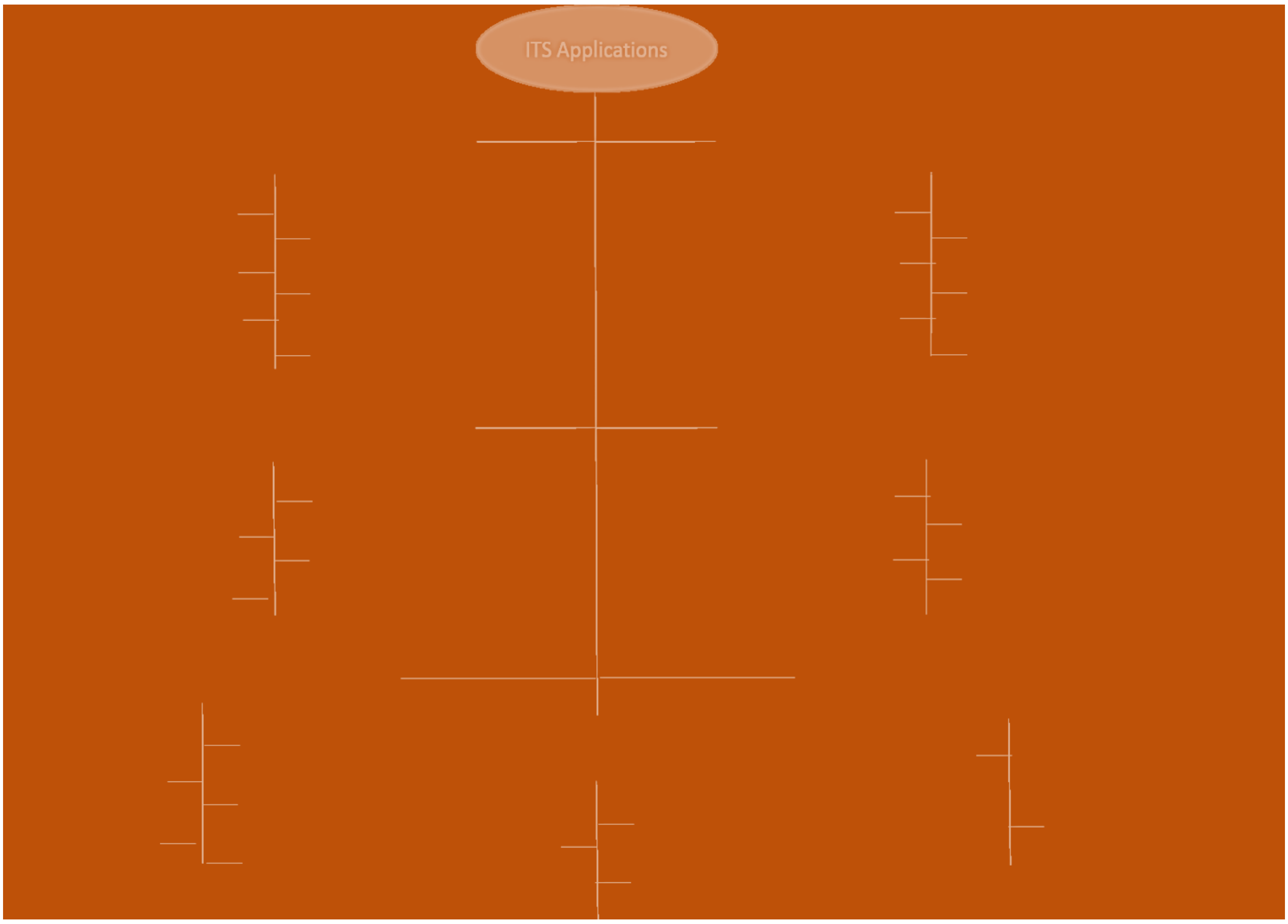

Figure 1: ITS applications

**A. Related Work**

ITS is an IOT application that aims to add technological advancement to conventional road transportation. In this perspective, we briefly introduce the studies that have been conducted in the field of context and context – awareness. In [15] Perera et al. surveyed the importance of context and context aware computing from an IoT perspective. They have evaluated nearly fifty projects to understand the challenges in the field of context – aware computing. The authors have also reviewed machine learning approaches in the reasoning phase of context – aware computing. In [16] Sezer et al. studied the importance of context in interactive applications where the user‘s context is changing rapidly. In [3] M. Veres and M. Moussa studied the importance of deep learning in intelligent transportation systems. The study also identified applications of deep learning within transportation. The authors have also identified open issues and challenges in the context of deep learning within ITS. In [17] L Zhui et al. surveyed the importance of big data analytics in intelligent transportation. The paper also proposed a framework of conducting big data analytics in ITS. Big data analytic methods such as supervised learning, unsupervised learning, reinforcement learning, deep learning and ontology based methods are discussed. In [18] H. V. Nejad et al. surveyed context aware computing for vehicular network applications. The authors have proposed a three layer classification framework consisting of environment, system-and-application, and context awareness. Based on their framework, they have reviewed and classified prominent ITS applications implemented in the real world.

In [19] proposed a framework for exploiting IOT for context – aware trust based personalized service. The authors have highlighted the need of user's preference in improving the quality of context aware recommendation systems.

The extensive study of these papers gives us an insight on exploring the use of context and context – awareness in intelligent transportation. The objective of this research paper is threefold: (i) to understand context and its various features in ITS (ii) to address the applicability of modelling techniques and reasoning approaches in ITS (iii) to shed light on impact of machine learning and IoT in ITS. The focus of this survey paper is towards understanding the importance of incurring context and reasoning in ITS application. It only highlights the machine learning approaches used in the reasoning phase. The detailed information on the design of machine learning algorithms is not covered in this article.

### B. Contribution

This research article is intended for researchers and developers who have profound interest in building context modelling and reasoning for ITS applications. The contribution of this article is summarized as follows:

- The major application areas of ITS are identified and classified.
- We identify the key features of context and its types to build context-aware intelligent transportation system
- We review the primary and secondary context used in prominent applications developed concerning to context-aware in ITS
- Compared to some related work in the literature, we review the state-of-the-art context modelling requirement for ITS
- We review a wide range of ITS applications that have used machine learning approach in context reasoning

The rest of the paper is organized as follows: Section II provides an introduction to context and context – awareness in ITS by discussing its various features and types. Section III focuses on context acquisition, modelling and reasoning techniques used in ITS applications. A succinct review of several modelling techniques and reasoning approaches used over the years is also presented. A discussion on the impact of IoT and machine learning in building context-aware systems for ITS application is presented in Section IV. The article is concluded in section V with summary of its main take away messages.

## II. Context-awareness in ITS

### A. Context and Context – awareness

The term context and context – awareness have been in literature since the 1990's and have evolved from desktop application, web computing, mobile computing, ubiquitous computing , cloud to IoT over the last decade. Context and context – aware grasped attention with the introduction of ubiquitous computing first coined by Schilit & Theimer. Since then, it has become a well-known research area in computer science. Many researchers have defined and proposed different explanations for context and context – awareness. We highlight some of those proposed by researchers. In [4] Schilit & Theimer define, 'context as information extracted from the environmental entities such as location, people, objects and changes to those objects'. In [5] Dey highlight, 'context as any information that can be used to characterise the situation of an entity. An entity being the user, location, date & time, and object that are relevant to the user or application. In [6] D. Abowd et al. define, 'context is information that describes the situation of an entity'. In [7] Brown et al. refer, 'context as entities such as location, time and season'. In [8] Ryan et al. define, 'context as user's location, identity, environment and time'. Franklin & Flaschbat [9] define, 'context as situation of the user'. Ward et al. [10] enumerate, 'context as state of the application'.

All these definitions though used in different research perspectives, one aspect remained fairly unchanged: they all define context (ex. People, location, time etc.) to be constantly changing the execution environment.

Context – aware in general defines the decision carried out by the system with relevance to the context. Context – aware is also used as a synonym with terms like adaptive and reactive. Here, we highlight some of the definitions proposed in the literature by researchers. Hull et al. [13] define, 'context – aware as ability of the system to detect, interpret and react to environment'. In [5] Dey highlights, 'context – aware as adaptation to the automation of a system based on the user's context'. Salber et al. [11] define, 'Context – aware as ability of the system to offer computational services based on the context'. In [12] Ward A et al. define, 'Context – aware systems are systems that dynamically change or adapt their behaviour based on the context of the system and the user'. In [8] Ryan et al. refers, 'context – aware applications to be applications that allow users to select context from the environment according to their interest'. In [7] Brown et al. define, 'context – awareness as the ability of the system to automatically react according to the context sensed by sensors'. In [6] D. Abowd et al. define, 'context – awareness as the use of context to provide tasks – relevant information to a user'.

These definitions on context – awareness relies on the key 'context' and to what kind of application the system is designed for. With reference to all these definitions, we incur that context – awareness is the ability of the system or application to relevantly respond to constantly changing execution environments. Intelligent transportation is also a constantly changing environment wherein any object/vehicle can communicate with any other object at any time and everywhere. Object communication in ITS generates vast amounts of data [14]. The major contribution of data generation in ITS application is through sensor technology and it plays a vital role in large – scale deployment of ITS. Nowadays, sensors come in less cost and small size yet good in high processing. This has led to an unpredictable range of data generation in a continuous environment. However, to add raw value to sensor data we have to understand the profound insight on it. The data will not have any value unless we analyze, interpret and understand it. The best way to understand the value of raw sensor data is through context and to apply context aware computing so the interpretation can be made easy.

The aim of this survey is to shed light on the role of context and context awareness has to play within ITS application. In particular, we focus on what is context in ITS, its types, and how different techniques are used in intelligent transportation for context data acquisition, context modelling and reasoning.

Many applications in ITS need to handle an extensive volume of data to provide on spot decisions to tackle traffic related problems. Using traditional data processing techniques to derive insight on data cannot happen in ITS where data is in the range of a trillion byte to petabyte. This is an at most challenge in ITS. Adding value to raw sensor data using context ensures fixing this issue and provides effective service delivery. ITS data can be obtained from diverse sources such as on-board terminals and sensors, GPS, video camera, base station, smart cards, web services, social media and so on. Understanding data and acquiring, modelling, reasoning context is vital in ITS application.

## B. Context Features

In [6] identified three context features that a context aware application can support. They are presentation, execution and annotation. We also identify an important feature a context – aware should support: dynamic configuration. All these features can be applied to ITS to sense the environment and context, and adapt the behaviour accordingly. Also, these features help in identifying different contexts with respect to ITS application scenarios.

**Presentation** – This focuses on what services to be delivered to the user. For example, every day when the user leaves from home to office, the system has to suggest the best route to reach office thereby avoiding traffic congestion. In this scenario, the context – aware applications need to acquire GPS data, model, analyze it and present it to the user. This uses context such as location, time, and user's identity.

**Execution** – ITS is both Human to Machine and more Machine to Machine communication. Sensors employed in the transportation object communicate with other sensor objects in the environment. Automating this communication among objects is a critical task in ITS. Let us consider an example when a user starts driving to a public mall, the system should not only show him the shortest and best path to travel but also automatically book a parking slot based on the context.

**Annotation** – Sensor data from a single sensor cannot provide enough information that can be used to infer high level inference. Therefore, data can be collected, analysed and fused together from multiple sensors. Annotation is also called context tagging. Let us take a traffic flow prediction scenario and discuss context annotation. To predict traffic flow on a particular day at a peak hour, sensor data from GPS, data from video cameras mounted on the road are fused together to analyze where the traffic density is high and low.

**Dynamic Configuration** – The system has to dynamically configure multiple functionalities depending on the application. For example, in traffic signal management, when an emergency vehicle like an ambulance arrives and if the lane is with red signal light, the system has to dynamically configure hardware functionalities by changing the red signal light to green signal for the ambulance to pass the lane.

## C. Context Types

Many researchers have identified different types of context based on different perspectives. In [15] C. Perera et al. analysed various context types and categorization. The authors have surveyed sixteen research papers and found different context types and scope. Further, they have identified the relationship between different context categories. In this paper we highlight the two important context types that can be applied in ITS.

**Primary context -** Any information gathered from location, identity, time, etc., without using existing context and data fusion technique is called primary context. Ex: Vehicle number, Location it travels.

**Secondary context** – Any information computed with the help of primary context is called secondary context. Ex: Vehicle number as primary context and number of people in the vehicle as secondary context.

Table 1. Provides an overview on primary and secondary context used in the prominent research study concerning context aware in ITS.

**Table 1**

**Review on Primary and Secondary context**

| **Year** | **Reference** | **Context – aware system** | **ITS Application** | **Primary Context** | **Secondary Context** |
|---|---|---|---|---|---|
| **2006** | René Meier et.al [20] | A spatial model for ITS services that deliver value added contextual information from systems to users | Advanced Public Transportation System (APTS) | Location<br>Time<br>Identity | Weather, Distance, Geographic shapes or regions |
| **2007** | Wolfgang Woerndl et al. [21] | Adding context value to VANET applications to improve performance. Presents a use case on context – aware gas station recommendation system. | Advanced Traveller Information system (ATIS) | Car ID<br>Location<br>Time | Acceleration, Gasoline level, Distance, Direction, Speed |
| **2008** | Simone Fuchs et al. [22] | Context aware system that assists and recommends driver, whether overtaking is wise or not | Advanced Vehicle Control System (AVCS | Driver<br>Traffic Signs<br>Time<br>Vehicle | Distance, Speed, Driver's skill, Road surface |
| **2009** | Jie SUN et al. [23] | A three layer context aware model for a smart car prototype | Advanced Vehicle Control System (AVCS) | Car<br><br>Driver<br><br>Environment | Engine status, Acclerograph, Gasoline, air bag, ABS<br><br>Physiological state of driver (ex:BP)<br><br>Weather, road signs, signal lamps |
| **2010** | Han-wen Chang et al. [24] | Taxi demand analysis by predicting hotspots using context | Commercial Vehicle Operation (CVO) | Time<br>Date, Taxi Location | Historical context, Weather, road name, landmarks, distance |
| **2011** | Vaninha Vieira et al. [25] | To access information about public transportation using it's context | Advanced Public Transportation System (APTS) | ID, location, route, time, station | speed , Meteorological conditions like weather, natural disasters |
| **2012** | Wael Alghamdi et al. [26] | Context aware driver assistance system to assist driver to avoid collision | Advanced Vehicle Control System (AVCS) | Driver, Vehicle, Location, Traffic sign | Speed, Driver's behaviour, Distance, blind spots |
| **2013** | M. V. Ramesh et al. [27] | Context aware system for real time road accident detection | Advanced Vehicle Control System (AVCS) | Vehicle Location Accident types | Landmark Severity of accident (ex: no of people injured) |
| **2014** | Yingling Fan et al. [28] | Android based smart phone to detect travel behaviour using context | Advanced Public Transportation System (APTS) | People, Location, Day/time | Activity, Travel behaviour (when, where, how) |
| **2015** | Chunzhao Guo et al. [29] | A vision based multimodal system for urban scenarios using road context | Advance Driver Assistance system (ADAS) | Vehicle, Location, Road/Lane | Geometry probability ( curvature of the lane), position |
| **2016** | M B Younes et al. [30] | Context – aware traffic light self-scheduling algorithm | Advanced Transport management System (ATMS) | Emergency vehicles (assigned priorities) | Traffic flow, traffic density |
| **2017** | Dennis Bohmlander et al. [31] | Context aware system to detect potential collisions and to activate safety actuators before accident occurs | Advanced Vehicle Control System (AVCS) | Vehicle, Environment, Driver, Location | Driving state, Position, turn rate, acceleration |
| **2018** | Razi Iqbal at al.[32] | Context based data analytics at fog level for Internet of Vehicles (IoV) | Advanced Vehicle Control System (AVCS) | Vehicle, Location, Route, User | Distance |
| **2019** | Suresh Chavhan et al. [33] | IoT-based Context-Aware Intelligent Public Transport System | Advanced Public Transportation System (APTS) | vehicles, staff, commuters, routes, Roadside Units (RSUs), Environment | Historical context information, Weather, Route condition (Traffic density), Vehicle condition |

## III. Context life cycle in ITS

Inferring contextual information from raw data requires context gathering, managing, evaluating and distributing it. Perara et al. [15] identified generic context life cycle phases for IoT application. We use this life cycle in ITS to explore context data acquisition, context modelling, context reasoning and context distribution. Figure 2. Illustrate this process.

### A. Context Acquisition

ITS uses diverse methods to acquire contextual information. In general, context in ITS is acquired through sensors/on-board terminals, cloud/fog and web services, environment infrastructure, and user interface.

**(i) Sensor/on – board terminals:** The context information of vehicle and user rely on sensors mounted in the vehicle and as well as any other sensors carried within the vehicle like mobile phones or smart bands on the driver's hand. Table 2 highlights different contexts acquired using on-board sensors.

**Table 2**
**Context data acquisition using sensors**

| On – board Sensor | Primary Context | Secondary Context |
|---|---|---|
| Vehicle Context | Vehicle, Location, type of vehicle, mass and weight | Vehicle direction, acceleration, longitudinal distance between neighbouring vehicles, humidity, friction, vehicle retardation, and vehicle's stopping distance, vehicle arrival time, vehicle departure time. |
| User Context | Driver identity, presence of pedestrian on lane | Driver's skill, behaviour, state of driver's eye, alcohol level in blood, heart rate, bp, human activities like walking, running, driving. |

**(ii) Cloud/fog and Web services**

Contextual information that remains unchanged for a quite long time can be acquired from cloud/internet and web services [55]. For example, the length, width and fuel type of vehicle can be acquired from a vehicle database stored in the cloud. Also, secondary context like weather, traffic congestion can also be acquired from cloud and web services.

**(iii) Environment Static Infrastructure**

Contextual information can also be acquired using environment infrastructure. It includes data server, central server or information systems. Bus schedule info systems in the bus stops, fuel price or gasoline rate provided in petrol banks, and parking policies are few examples of static infrastructure.

**(iv) User Interface**

User interfaces used by users allow contextual data acquisition in ITS. Usually, users make use of interfaces like smart phones or handheld smart devices to give input to the systems. For example, in

on – street parking applications, parking preferences and destinations are entered by the user. Similarly, in the road – accident system, the number of vehicles damaged and collided, number of people injured, current state of collision are provided by people around the vicinity of the accident scene.

## B. Context Modelling

In ITS, data gathered from a variety of sources differ in the quality of the information they produce. Therefore, context representation is required to understand the properties, features and details of the context. There are different types of context modelling techniques in the literature [16] [34] such as key-value modelling, mark up scheme modelling, graphical modelling, object oriented modelling, logic based modelling and ontology based modelling. In [15] the functionality of these context models is best explained. We here present the applicability of these models from ITS perspective. Table 3 highlights the context modelling requirement for ITS and applicability of the models.

(i) **Key – value modelling** – One of the oldest approaches in context modelling is key – value modelling. It is the simplest form of representation technique that models key-value pairs in format such as text files. This model does not support data retrieval and reasoning. This model is not suitable for ITS since it can be used to model only a limited amount of data.

(ii) **Mark-up modelling** – Mark-up modelling is more structured when compared to key-value models. It models context within tags and uses mark-up schemes like XML and JSON to store data, transfer data among applications. It allows efficient data retrieval only with fewer levels of information processing and does not provide effective validation support. There are many levels of information processing involved in ITS application. For example, in route prediction, sensor data from GPS and images from camera sensors are both processed and analysed to predict optimal routes. Therefore, the use of mark - up modelling to retrieve information in an application having many levels of information is difficult.

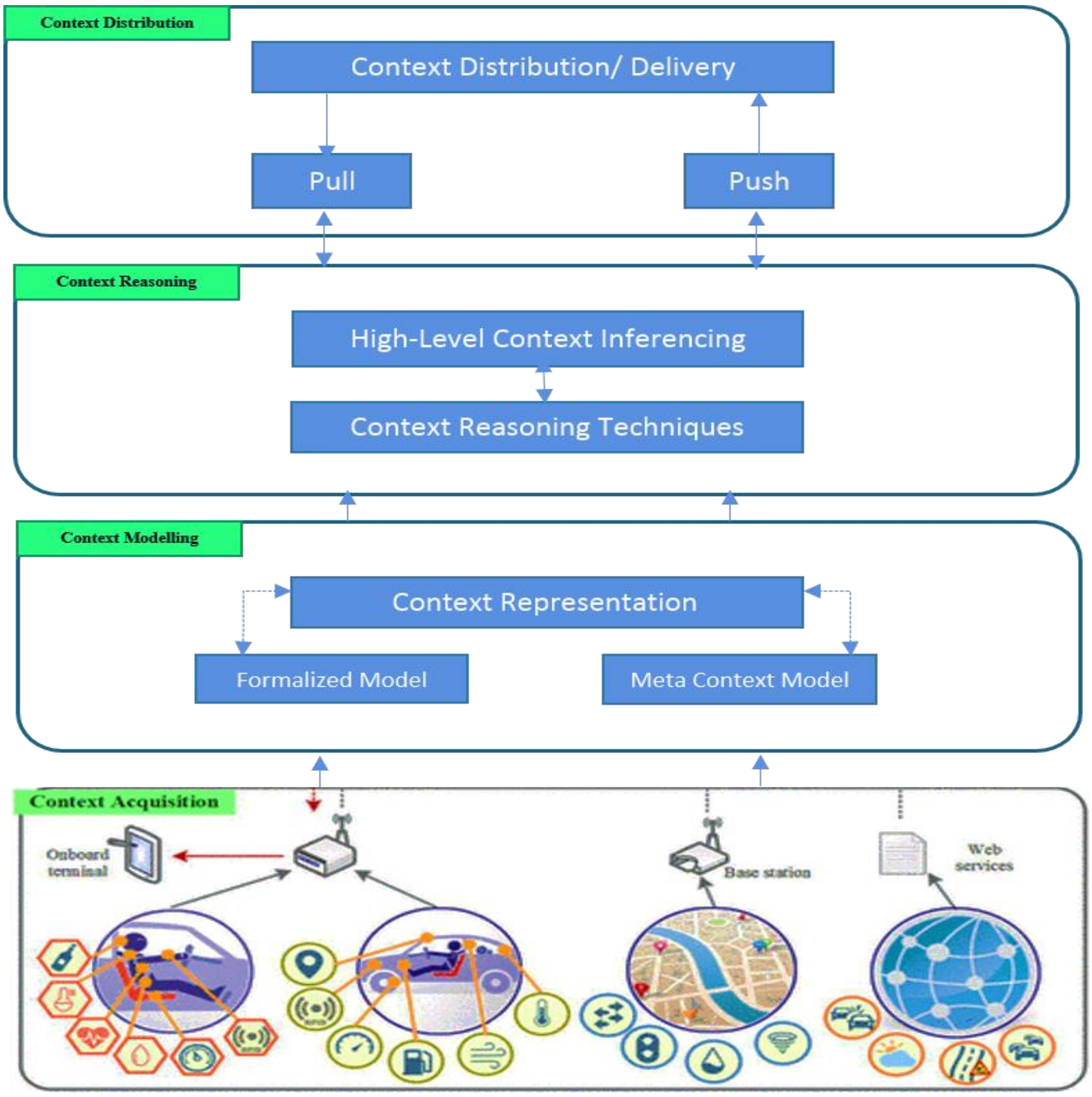

Figure 2. Context Life Cycle

(iii) **Graphic based modelling** – ITS application involves quick manipulation and decision making on the fly. This model will help in context runtime manipulation and allows relationship modelling. It represents context with relationships. Unified Modelling Language and Object Role Modelling Language are object based modelling techniques that allow relationships to be captured into context model.  Relational dbs and NoSQL dbs can also be used as graphical modelling techniques in ITS application development.

(iv) **Object based modelling** – It models context as class hierarchy and relationship. It supports all the features of object oriented paradigm such as encapsulation, abstraction and reusability. Most of the ITS application development uses high level programming and therefore modelling the context as hierarchy and integrating it into the application can be done easily.

(v) **Logic based modelling** – In this model context is represented as rules and facts. This model is used to generate high-level context using low level-context. Logical model supports reasoning to a certain level. Therefore, it has the capability to enhance other modelling techniques. In ITS, modelling context based on logical rules and expressions helps to predict accident occurrence and event detection.

(vi) **Ontology based modelling** – To add value to raw sensor data and to model domain knowledge, context representation in ITS uses ontology modelling. This model supports high semantic reasoning and allows relationship modelling. According to the studies in the literature, in context aware computing and in applications where sensor data generation is more, ontologies are the preferred modelling technique. In this model, context is organized into ontologies using semantic web ontology languages like RDF, RDFS and OWL. Ontology is the main component of semantic web technology used for both modelling and reasoning. Ontologies support high level context inferring, separate domain knowledge from operational knowledge and hence enable reuse of domain knowledge to be integrated with other domains in application development.

(vii) **Spatial model** – This model can be fact or object or ontology based and are suited for applications that are mainly location based. Space is an important context in the transportation system. The spatial entity in an ITS application can denote a place where the object is, and resources around the object. Thus, spatial modelling can be used for ITS applications that organize their information by physical location. This physical location can be (i) geometric (for example, latitude and longitude of physical location sensed by GPS) (ii) symbolic (for example, address and landmarks)

(viii) **Hybrid Model** – Hybrid model is the combination of any other modelling technique. It integrates the best features of other models to represent the context.

**Summary of Context modelling** – From Table 3 we can infer that no single model all alone will be able to satisfy the identified requirements for ITS. For example, a Key-value model for predicting traffic flow application is not a scalable model to store complex data structure. Further, it cannot be used to represent relationships between contexts like traffic density and peak hours. This two context information has to be captured to ensure correct behaviour of the application. Whereas, using ontology modelling for the same application is scalable as it supports complex data structure and enables relationship representation. ITS application, which involves data generation based on spatial context and sensor devices can be represented using spatial modelling. Table 4 presents a review on modelling techniques used in prominent ITS research applications developed in the recent past.

**Table 3. Context modelling requirement for ITS**

| Modelling technique | Context modelled as | High volume of data generation | Relationship representation | Semantic/ Logical reasoning | Information retrieval | Validation support |
|---|---|---|---|---|---|---|
| **Key – Value** | Attribute - value | No support | Not possible | No support | Hard | No |
| **Mark up** | Tags | No Support | Not Possible | No support | Hard | Yes |
| **Graphical** | Relations | Support | Possible | Moderate support | Moderate | Yes |
| **Object** | class | Support | Possible | Moderate support | Hard | No |
| **Logic based** | Rules and facts | Support | Possible | Moderate support | Hard | No |
| **Ontology based** | Ontologies | Support | Possible | High support | Hard | Yes |

## C. Context Reasoning

Context reasoning is otherwise called inferencing and can be described as inferring unknown state from the known state. It can also be defined as extracting new knowledge from the available context. The main purpose of context reasoning in ITS is to address the imperfection and ambiguity present in the data. There are three important steps involved in context reasoning: Context pre-processing, sensor fusion and context inference [35]. In the first step, data is cleansed to define relevant context attributes in the raw data. It uses mining techniques like feature extraction, dimensionality reduction to remove outliers and to fill the missing data. In the second step, sensor data from multiple sensors are fused together to retrieve for more accurate, complete and reliable data. This step is vital in ITS application as a huge amount of data gets generated with the use of sensors. The third step, context inference, infer new knowledge/ high-level context information from low-level context. There are many context reasoning techniques present in the literature. We categorize this as fuzzy-based, ontology-based, probabilistic-based, rule-based and machine learning-based techniques.

(i) **Fuzzy logic** – Fuzzy logic supports approximate reasoning instead of fixed reasoning. In fuzzy, apart from 0 and 1, partial truth is also represented. It uses natural definition rather than numerical value. For example, instead of mentioning the speed of a vehicle as 160 km/hr, it can be represented in natural form as speed: very fast. Several examples for fuzzy logic to represent context are addressed in [36].

(ii) **Ontology-based** – This model uses description logic for context reasoning. One advantage of ontology reasoning is it can be well integrated with ontology modelling. This method is used widely in applications such as agriculture, and medicinal science, where domain knowledge is essential. This technique is also used for sensor data fusion in sensor communication [37]. However, the disadvantage is, it is difficult to integrate ontologies with context pre-processing steps to find the missing values in the sensor data. Fig 3 depicts the ontology for three contexts namely car, person and environment in ITS.

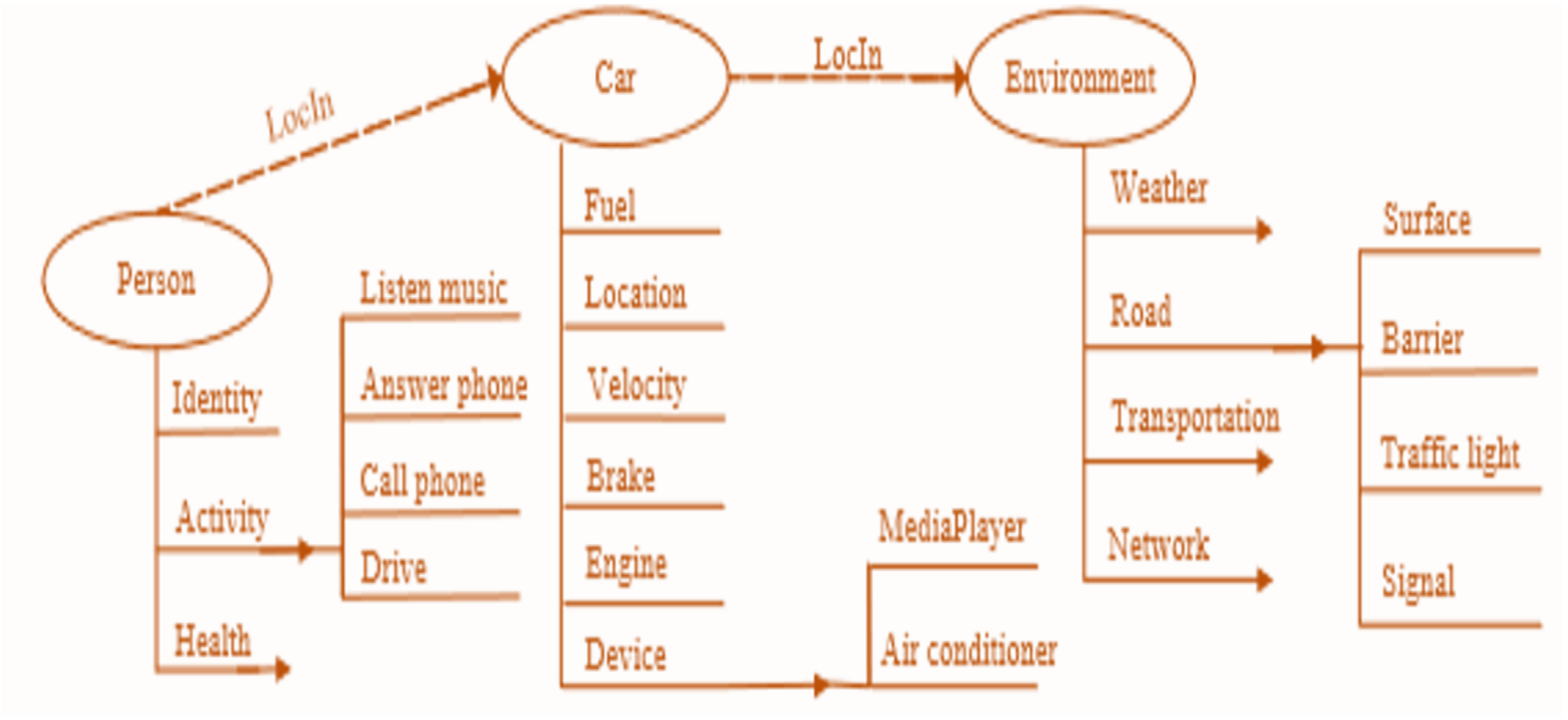

Figure 3. OntOntology for car, Person and Environment

(iii) **Probabilistic model** – This reasoning technique allows decision to be made based on occurrence of event probabilities. Mainly used in ITS applications to predict the next event, and to recognize activities of pedestrians like walking, running, cycling etc. Hidden Markov Model [38] is used as a probabilistic technique in ITS application to bridge the gap between low-level context data such as GPS sensor and high-level context information such as mode of transportation and user destination.

(iv) **Machine learning model** – In recent years, due to the increase in growth of data, machine learning reasoning techniques are used widely in many applications. In ITS, the use of machine learning technique is used to a great extent to infer new knowledge from the low-level context. There is another class of ML technique called deep learning technique which is also used to infer context information. DL techniques are widely applied to infer knowledge from image data. In both ML and DL there are three types of learning called supervised learning, unsupervised learning, and reinforcement learning.

**Supervised Learning** – It is a class of learning technique, where the trained data is used to make the machine learn from labelled data to predict the accurate output for the test data. This technique is used in ITS applications to classify the objects. For example, in [39] uses supervised learning for semantic segmentation for autonomous vehicle, this technique is used to classify objects in the road vicinity such as vehicles, pedestrians, road signs, environment aspects like tree, sky etc., There are many classes of supervised algorithms like ANN, decision tree, Bayesian network, and support vector machines. In [40] an ELM based classification technique is proposed to detect licence plates from vehicle images.

**Unsupervised Learning** – It is a reasoning technique that is used to extract hidden information from unlabelled data. K-Nearest neighbour cluster technique is widely used for context-aware reasoning. This technique is used in ITS to precisely cluster types of vehicles, spots and locations having similarities. ANN self-organizing maps is used in [41] for classifying sensor data for context-aware applications in ITS.

**Reinforcement Learning** – It is a class of unsupervised techniques which learns from the environment. A reward or feedback signal helps the system to reason more and its behaviour. Hence it is called reinforcement learning. RL in ITS applications is used to improve the system performance. Q-Learning is widely used in context-aware applications of ITS [42][57][58].

**Summary of Context reasoning** – In Table 4, we have presented a review on context reasoning techniques widely used in the applications of ITS. Each of these techniques can be applied in many different applications of ITS like destination prediction, travel time estimation, demand serving, traffic signal control etc. The combination of these techniques as hybrid reasoning can also be used to reduce the weakness, and thus complementing each other. It is recommended to use ontology reasoning technique with ontology modelling to incur more meaningful context. It is also found, in recent years, researchers have shown interest to adapt machine learning based reasoning techniques to incur context knowledge.

## D. Context Distribution

The last phase in context life cycle is context distribution which delivers context to the users. Context acquisition is again context distribution from the user's perspective. This can happen in two ways. One, the user can periodically change his query according to his usage. Second, if the user subscribes to the system, the context system periodically outputs specific sensor data when an event occurs. Hence, the context life cycle process is cyclic. Where, new context is modelled in the modelling phase and high-level context is inferred from low-level context in the reasoning phase. And, the new context is delivered to the user.

## IV. Discussion

It is clear from the aforementioned sections, huge amounts of data are generated using various sensors in ITS application. These sensors continuously sense and generate enormous amounts of data of which most are raw data, in fact highly irregular and heterogeneous. To add value to this raw data, we need to analyse, interpret and understand it. Context-aware computing has proven to be the best in the literature to model, reason and distribute the underlying value of data as context. Undoubtedly, sensor technology and Internet of Things (IoT) plays a significant role in enabling ITS development. The impact of IoT on ITS is extensive. Ranging from wireless sensor technology to communication protocols like Wi-Fi, RFID, GPS etc., use of data fusion to combine variety of sources, middleware solution to collect, process and analyse context, to enable security and privacy, IoT handshakes with WSN, cloud and semantic technologies to make conventional transportation more convenient, easy and accessible.

Similarly, machine learning and emerging deep learning approaches have great impact in adding intelligence to conventional transportation. These approaches have emerged to help overcome many of the challenges faced in ITS like removal of noisy data and data imputation, optimizing scarcity in standardized environments, collecting and annotating large amounts of data for training, learning for sustainable commuting, optimizing level of services and learning with multiple sensor models. Therefore, using an IoT architecture and ML approach to design context-aware ITS provide solutions to data processing, context discovery, modelling and reasoning, data fusion, resource sharing, real time processing, knowledge management, event detection and prediction. Yet many challenges exist for future research such as,

- Automated configuration of sensors
- Incorporating and integrating multiple modelling and reasoning techniques together (to address where to employ which technique)
- Support for middleware to provide Quality of Services (e.g.: heterogeneity, interoperability and dependability)
- Security and privacy- to enable security mechanism to avoid misuse of context at ITS application level and middleware level
- To enable trust among users – devising a trust based policy where the acceptance of the user is essential

## V. Conclusion

ITS is a rapidly changing environment, where decisions are to be taken on the fly to solve transportation problems to make it more safer, better informed, and coordinated. Understanding raw sensor data in ITS is one of the biggest challenges. To address this challenge, the paper focused on elaborating and identifying context features and context types. In this survey paper, we reviewed context modelling and context reasoning techniques and its applicability is summarized. It is clear from the survey, the more data acquisition in ITS is through sensors and to infer semantics from raw sensor data, context aware computing uses modelling and reasoning techniques. It is also identified, in recent years, machine learning and deep learning techniques are widely used in the reasoning phase of ITS. The impact of IoT and machine learning in the development of ITS application is discussed and future challenges are identified.

**Table 4 Context modelling and Context Reasoning Techniques**

| Year | Reference | ITS Application | Context Modelling Technique | Context Reasoning Technique |
|---|---|---|---|---|
| **2015** | Chunzhao Guo et al. [29] | A vision based multimodal system for urban scenarios using road context | Knowledge/ontology based modelling | Machine learning (HMM model for road detection) |
| | De. Brebissin et al. [43] | Destination prediction | Spatial modelling | Supervised – Backpropagation model |
| | Lv et al. [44] | Traffic flow prediction | Knowledge based Historical traffic flow | Supervised – SAE model |
| **2016** | Toque et al. [45] | Demand prediction | Metadata model | Supervised – Long Short Term Memory model |
| | Siripanpornchana [46] | Travel time estimation | Knowledge based/ historical travel time | Supervised - DBN |
| **2017** | Dennis Bohmlander et al. [31] | Context aware system to detect potential collisions and to activate safety actuators before accident occurs | Knowledge database modelling | Machine learning (Supervised – Support vector machine) |
| | Xu et al. [47] | Demand prediction | Knowledge based | Supervised learning – Long Short Term Memory |
| | Jiana et al. [48] | Transportation mode prediction | Spatial Modelling | Supervised – Bidirectional GRU |
| **2018** | Yao et al. [49] | Demand prediction | Hybrid model | Supervised learning – Backpropagation model |
| | Jind et al. [50] | Demand serving | Spatial modelling | Supervised learning and Reinforcement learning |
| | Nishi et al.[51] | Traffic signal control | - | Reinforcement Learning |
| | Krishnakumari [52] | Traffic state classification | Spatial modelling | Deep learning – Convolutional neural network |
| **2019** | Suresh Chavhan et al. [33] | Intelligent Public Transport System | Probabilistic model | ML with agent techniques. |
| | Gang et al. [53] | Demand prediction | Knowledge based / Historical data | Supervised learning – Recurrent neural network |
| | Liang et al. [54] | Traffic signal control | Knowledge base | Reinforcement learning |